\documentclass[journal, twocolumn]{IEEEtran}
\usepackage[algo2e]{algorithm2e}
\usepackage{algorithm}
\usepackage{amsthm,amssymb,amsmath,graphicx,paralist,subfigure,pdfpages,cite,url}
 
\newtheorem{assump}{Assumption}

\def\ln{{\rm ln}}

\def\mc{\mathcal}
\def\mb{\mathbf}
\def\mbb{\mathbb}
\def\ra{\rightarrow}
\IEEEoverridecommandlockouts

\def\mbb{\mathbb}
\def\mb{\mathbf}
\def\mc{\mathcal}

\def\wt{\widetilde}

\begin{document}
\title{Distributed Nesterov gradient methods\\ over arbitrary graphs} 
\author{Ran Xin, Du\v{s}an Jakoveti\'{c}, and Usman A. Khan\thanks{
RX and UAK are with the Department of Electrical and Computer Engineering, Tufts University, USA. {\texttt{\{ran.xin@,khan@ece.\}tufts.edu}} 

DJ is with the Department of Mathematics and Informatics, Faculty of Science, University of Novi Sad, Serbia. {\texttt{djakovet@uns.ac.rs}} }}

\maketitle
\begin{abstract}
In this letter, we introduce a distributed Nesterov method, termed as~$\mc{ABN}$, that does not require doubly-stochastic weight matrices. Instead, the implementation is based on a simultaneous application of both row- and column-stochastic weights that makes this method applicable to arbitrary (strongly-connected) graphs. Since constructing column-stochastic weights needs additional information (the number of outgoing neighbors at each agent), not available in certain communication protocols, we derive a variation, termed as \textit{FROZEN}, that only requires row-stochastic weights but at the expense of additional iterations for eigenvector learning. We numerically study these algorithms for various objective functions and network parameters and show that the proposed distributed Nesterov methods achieve acceleration compared to the current state-of-the-art methods for distributed optimization.  
\end{abstract}

\section{Introduction}\label{s1}
Distributed optimization has recently seen a surge of interest particularly with the emergence of modern signal processing and machine learning applications. A well-studied problem in this domain is finite sum minimization that also has some relevance to empirical risk formulations, i.e., 
\begin{align*}
\min_{\mb{x}} \sum_{i} f_i(\mb{x}),
\end{align*}
where each~$f_i:\mbb{R}^p\ra\mbb{R}$ is a smooth and convex function available at an agent~$i$. Since the~$f_i$'s depend on data that may be private to each agent and communicating large data is impractical, developing distributed solutions of  the above problem have attracted a strong interest. Related work has been a topic of significant research in the areas of signal processing and control~\cite{distributed_Rabbit,distributed_Jianshu,ESOM,safavi2018distributed}, and more recently has also found coverage in the machine learning literature~\cite{forero2010consensus,distributed_Boyd,wai2018multi,DGD_nips,hong2017prox,scaman2018optimal}. 

Since the focus is on distributed implementation, the information exchange mechanism among the agents becomes a key ingredient of the solutions. Such inter-agent information exchange is modeled by a graph and significant work has focused on algorithm design under various graph topologies. The associated algorithms require two key steps: (i) consensus, i.e., reaching agreement among the agents; and, (ii) optimality, i.e., showing that the agreement is on the optimal solution. Naturally, consensus algorithms have been predominantly used as the basic building block of distributed optimization on top of which a gradient correction is added to steer the agreement to the optimal solution. Initial work thus follows closely the progress achieved in the consensus algorithms and extensions to various graph topologies, see e.g.,~\cite{cc_Ram,uc_Nedic,Angelia_Async,forero2010consensus,distributed_Boyd,cc_Lee,sam_spl1:15}.

Early work on consensus assumes doubly-stochastic (DS) weights~\cite{c_Saber,xiao2004fast}, which require the underlying graphs to be undirected (or balanced) since both incoming~and outgoing weights must sum to~$1$. The subsequent~work~on~optimization over undirected graphs includes~\cite{uc_Nedic} where the convergence is sublinear and~\cite{AugDGM,harness,EXTRA} with linear convergence. For directed (and unbalanced) graphs, it is not possible to construct DS weights, i.e., the weights can be chosen such that they sum to~$1$ \textit{either} only on incoming edges \textit{or} only on outgoing edges. Optimization over digraphs~\cite{opdirect_Tsianous2,opdirect_Nedic,D-DPS,D-DGD,add-opt,diging,linear_row,FROST} thus has been built on consensus with non-DS weights~\cite{ac_directed0,ac_directed,ac_Cai1}. Required now is a division with additional iterates that learn the non-$\mb{1}$ (where~$\mb{1}$ is a vector of all~$1$'s) Perron eigenvector of the underlying weight matrix, see~\cite{D-DPS,add-opt,diging} for details. Such division causes significant conservatism and stability issues~\cite{ABM}.

Recently, we introduced the~$\mc{AB}$ algorithm that removes the need of eigenvector learning by utilizing both row-stochastic (RS) and column-stochastic (CS) weights, simultaneously,~\cite{AB}. The algorithm thus is applicable to arbitrary strongly-connected graphs. The intuition behind using both sets of weights is as follows: Let~$A$ be RS and~$B$~be CS, with~$\mb{w}^\top A=\mb{w}^\top$ and~$B\mb{v}=\mb{v}$, in addition to being primitive. From Perron-Frobenius theorem, we have that~$A^\infty=\mb{1}\mb{w}^\top$ and~$B^\infty=\mb{v}\mb{1}^\top$. Clearly, using~$A$ or~$B$~alone~makes~an~algorithm dependent on the non-$\mb{1}$ Perron eigenvector~($\mb{w}$ or~$\mb{v}$) and thus the need for the aforementioned division by the iterates learning this eigenvector. Using~$A$ and~$B$ simultaneously, the asymptotics of~$\mc{AB}$ are driven by, loosely speaking,~$A^\infty B^\infty=(\mb{w}^\top\mb{v})\cdot\mb{1}\mb{1}^\top$, which recovers the consensus matrix,~$\mb{1}\mb{1}^\top$, without any scaling. It is shown in~\cite{AB} that~$\mc{AB}$ converges linearly to the optimal for smooth and strongly-convex functions. 

In this letter, we study accelerated optimization over arbitrary graphs by extending~$\mc{AB}$ with Nesterov's momentum. We first propose~$\mc{ABN}$ that uses both RS and CS weights. Construct CS weights requires each agent to know at least its out-degree, which may not be possible in broadcast-type communication scenarios. To address this challenge, we provide an alternate algorithm, termed as \textit{FROZEN}, that only uses RS weights. We show that \textit{FROZEN} can be derived from~$\mc{ABN}$ with the help of a simple state transformation. Finally, we note that a rigorous theoretical analysis is beyond the scope of this letter and we present extensive simulations to highlight and verify different aspects of the proposed methods.
 
We now describe the rest of this paper. Section~\ref{s2} formulates the problem and recaps the~$\mc{AB}$ algorithm. Section~\ref{s3}~describes the two methods,~$\mc{ABN}$ and \textit{FROZEN}, and Section~\ref{s4} provides simulations comparing the proposed methods with the state-of-the-art in distributed optimization over both convex and strongly-convex functions, and over various digraphs. 

\section{Problem Formulation and Preliminaries}\label{s2}
Consider~$n$ agents connected over a digraph,~$\mc{G}=(\mc{V},\mc{E})$, where~$\mc{V}=\{1,\cdots,n\}$ is the set of agents and~$\mc{E}$ is the~collection of edges,~$(i,j),i,j\in\mc{V}$, such that~$j\rightarrow i$. We define~$\mc{N}_i^{{\scriptsize \mbox{in}}}$ as the collection of in-neighbors of agent~$i$, i.e., the set of agents that can send information to agent~$i$. Similarly,~$\mc{N}_i^{{\scriptsize \mbox{out}}}$ is the set of out-neighbors of agent~$i$. Note that both~$\mc{N}_i^{{\scriptsize \mbox{in}}}$ and~$\mc{N}_i^{{\scriptsize \mbox{out}}}$ include node~$i$. The agents solve the following  unconstrained optimization problem:
\begin{align}
	\mbox{P1}:
	\quad\min_{\mb{x}\in\mathbb{R}^p}F(\mb{x})\triangleq\frac{1}{n}\sum_{i=1}^nf_i(\mb{x}),\nonumber
\end{align}
where each~$f_i:\mbb{R}^p\rightarrow\mbb{R}$ is private to agent~$i$. We formalize the set of assumptions as follows.
\begin{assump}\label{asp1}
	The  graph,~$\mc{G}$, is strongly-connected.
\end{assump}

\begin{assump}\label{asp2}
	Each local objective,~$f_i$, is~$\mu$-strongly-convex,~$\mu>0$, i.e.,~$\forall i\in\mc{V}$ and~$\forall\mb{x}, \mb{y}\in\mbb{R}^p$, we have
	\begin{equation*}
		f_i(\mb{y})\geq f_i(\mb{x})+\nabla f_i(\mb{x})^\top(\mb{y}-\mb{x})+\frac{\mu}{2}\|\mb{x}-\mb{y}\|^2.
	\end{equation*}
\end{assump} 

\begin{assump}\label{asp3}
	Each local objective,~$f_i$, is~$L$-smooth, i.e., its gradient is Lipschitz-continuous:~$\forall i\in\mc{V}$ and~$\forall\mb{x}, \mb{y}\in\mbb{R}^p$, we have, for some~$L>0$,
	\begin{equation*}
		\qquad\|\mb{\nabla} f_i(\mb{x})-\mb{\nabla} f_i(\mb{y})\|\leq L\|\mb{x}-\mb{y}\|.
	\end{equation*}
\end{assump}
Let~$\mc{F}_{L}^{1,1}$ be the class of functions satisfying Assumption~\ref{asp3} and let~$\mc{F}_{\mu,L}^{1,1}$ be the class of functions that satisfy both Assumptions~\ref{asp2} and~\ref{asp3}; note that~$\mu\leq L$. In this letter, we propose distributed algorithms to solve Problem P1 for both function classes, i.e.,~$F\in\mc{F}_{L}^{1,1}$ and~$F\in\mc{F}_{\mu,L}^{1,1}$. We assume that the underlying optimization is solvable in the class~$\mc{F}_{L}^{1,1}$.

\vspace{-0.1cm}
\subsection{Centralized Optimization: Nesterov's Method}
The gradient descent algorithm is given by
\begin{equation*}
\mb{x}_{k+1} = \mb{x}_{k} - \alpha\nabla F\left(\mb{x}_{k}\right), 
\end{equation*}
where~$k$ is the iteration and~$\alpha$ is the step-size. It is well known~\cite{polyak1987introduction,nesterov2013introductory} that the oracle complexity of this method to achieve an~$\epsilon$-accuracy is~$\mc{O}(\frac{1}{\epsilon})$ for the function class~$\mc{F}_L^{1,1}$ and~$\mc{O}(\mc{Q}\log\frac{1}{\epsilon})$ for the function class~$\mc{F}_{\mu,L}^{1,1}$, where~$\mc{Q}\triangleq\tfrac{L}{\mu}$ is the condition number of the objective function,~$F$. There are gaps between the lower oracle complexity bounds of the function class~$\mc{F}_L^{1,1}$ and~$\mc{F}_{\mu,L}^{1,1}$, and the upper complexity bounds of gradient descent~\cite{nesterov2013introductory}. This gap is closed by the seminal work~\cite{nesterov2013introductory} by Nesterov, which accelerates the convergence of the gradient descent by adding a certain momentum to gradient descent. The centralized Nesterov's method~\cite{nesterov2013introductory} iteratively updates two variables~$\mb{x}_k,\mb{y}_k\in\mathbb{R}^p$, initialized arbitrarily with~$\mb{x}_0=\mb{y}_0$, as follows: 
 \begin{subequations}
\begin{align}\label{CN}
\mb{y}_{k+1} &= \mb{x}_{k} - \frac{1}{L}\nabla F(\mb{x}_{k}), \\ 
\mb{x}_{k+1} &= \mb{y}_{k+1} + \beta_k ( \mb{y}_{k+1}-\mb{y}_{k} ),
\end{align}
\end{subequations} 
where~$\beta_k$ is the momentum parameter. For the function class $\mc{F}_L^{1,1}$, choosing~$\beta_k = \frac{k}{k+3}$ leads to an optimal oracle complexity of~$\mc{O}(\frac{1}{\sqrt{\epsilon}})$, while for the function class~$\mc{F}_{\mu,L}^{1,1}$,~$\beta_k = \frac{\sqrt{L}-\sqrt{\mu}}{\sqrt{L}+\sqrt{\mu}}$ results into an optimal oracle complexity of~$\mc{O}(\sqrt{\mc{Q}}\log\frac{1}{\epsilon})$. 

\subsection{Distributed Optimization: The~$\mc{AB}$ algorithm}\label{secAB}
When the objective functions are not available at a central location, distributed solutions are required to solve Problem P1. Most existing work~\cite{distributed_Rabbit,distributed_Jianshu,ESOM,cc_Ram,uc_Nedic,Angelia_Async,cc_Lee,AugDGM,harness,EXTRA} is restricted to undirected graphs, since the weights assigned to neighboring agents must be doubly-stochastic. The work on directed graphs~\cite{opdirect_Tsianous2,opdirect_Nedic,add-opt,diging,linear_row,FROST} is largely based on push-sum consensus~\cite{ac_directed0,ac_directed} that requires eigenvector learning. Recently,~$\mc{AB}$ algorithm was introduced in~\cite{AB} that does not require eigenvector learning by utilizing a novel approach to deal with the non-doubly-stochasticity in digraphs. 

We now describe the~$\mc{AB}$ algorithm: Consider two distinct sets of weights,~$\{a_{ij}\}$ and~$\{b_{ij}\}$, at each agent such that 
\begin{align*} 
a_{ij}&=\left\{
\begin{array}{rl}
>0,&j\in\mc{N}_i^{{\scriptsize \mbox{in}}},\\
0,&\mbox{otherwise},
\end{array}
\right.
\quad
\sum_{j=1}^na_{ij}=1,\forall i,
\end{align*} 
\begin{align*} 
b_{ij}&=\left\{
\begin{array}{rl}
>0,&i\in\mc{N}_j^{{\scriptsize \mbox{out}}},\\
0,&\mbox{otherwise},
\end{array}
\right.
\quad
\sum_{i=1}^nb_{ij}=1,\forall j. 
\end{align*}
In other words, the weight matrix,~$A=\{a_{ij}\}$, is row-stochastic, while~$B=\{b_{ij}\}$ is column-stochastic. It is straightforward to note that the construction of row-stochastic weights,~$A$, is trivial as it each agent~$i$ on its own assigns arbitrary weights to incoming information (from agents in~$\mc{N}_i^{{\scriptsize \mbox{in}}}$) such that these weights sum to~$1$. The construction of column-stochastic weights is more involved as it requires that all outgoing weights at agent~$i$ must sum to~$1$ and thus cannot be assigned on incoming information. The simplest way to obtain such weights is for each agent~$i$ to transmit~${\mb{s}_k^i}/{|\mc{N}_i^{{\scriptsize \mbox{out}}}|}$ to its outgoing neighbors in~$\mc{N}_i^{{\scriptsize \mbox{out}}}$. This strategy, however, requires the knowledge of the out-degree at each agent~$i$.

With the help of the row- and column-stochastic weights, we can now describe the~$\mc{AB}$ algorithm as follows~\cite{AB}:
 \begin{subequations}\label{AB}
 	\begin{align}
 	\mb{x}^i_{k+1}&=\sum_{j=1}^{n}a_{ij}\mb{x}^{j}_k-\alpha\mb{s}^i_k, 
 	\label{ABa}\\
 	\mb{s}^i_{k+1}&=\sum_{j=1}^{n}b_{ij}\mb{s}^j_k+\nabla f_i\big(\mb{x}^i_{k+1}\big)-\nabla f_i\big(\mb{x}^i_k\big), \label{ABb}
 	\end{align}
 \end{subequations}
where~$\mb{x}_0^i\in\mbb{R}^p$ is arbitrary and~$\mb{s}_0^i=\nabla f_i(\mb{x}_0^i)$. We explain the above algorithm in the following. Eq.~\eqref{ABa} essentially is gradient descent where the descent direction is~$\mb{s}_k^i$, instead of~$\nabla f_i(\mb{x}_k^i)$ as used in the earlier methods~\cite{uc_Nedic,D-DGD}. Eq.~\eqref{ABb}, on the other hand, is gradient tracking, i.e.,~$\mb{s}_k^i\ra\sum_i\nabla f_i(\mb x_k^i)$, and thus Eq.~$\eqref{ABa}$ descends in the global direction, asymptotically. It is shown in~\cite{AB} that~$\mc{AB}$ converges linearly to the optimal solution for the function class~$\mc{F}^{1,1}_{\mu,L}$. 

The~$\mc{AB}$ algorithm for undirected graphs where both weights are doubly-stochastic was studied earlier in~\cite{AugDGM,harness,diging}. It is shown in~\cite{harness} that the oracle complexity with doubly-stochastic weights is~$\mc{O}(Q^2 \log \frac{1}{\epsilon})$. Extensions of~$\mc{AB}$ include: non-coordinated step-sizes and heavy-ball momentum~\cite{ABM}; time-varying graphs~\cite{TVAB,TVpushpull}; analysis for non-convex functions~\cite{NCAB}. Related work on distributed Nesterov-type methods can be found in~\cite{fast_Gradient,Dusan_Ne,dnesterov}, which is restricted to undirected graphs. There is no prior work on Nesterov's method that is applicable to arbitrary strongly-connected graphs.
 
\newpage
\section{Distributed Nesterov Gradient Methods}\label{s3}
In this section, ww introduce two distributed Nesterov gradient methods, both of which are applicable to arbitrary, strongly-connected, graphs. 

\subsection{The~$\mc{ABN}$ algorithm}
Each agent,~$i\in\mc{V}$, maintains three variables:~$\mb{x}^{i}_k$,~$\mb{y}^{i}_k$ and~$\mb{s}_k^i$, all in~$\mbb{R}^p$, where~$\mb{x}^i_k$ and~$\mb{y}^i_k$ are the local estimates of the global minimizer and~$\mb{s}^i_k$ is used to track the average gradient. The~$\mc{ABN}$ algorithm is described in Algorithm~\ref{ABNalg}.
\begin{algorithm}[H]
At each agent~$i$:\\
\textbf{Initialize:} Arbitrary~$\mb{x}_0^i=\mb{y}_0^i\in\mbb{R}^p$ and~$\mb{s}_0^i=\nabla f_i(\mb{x}_0^i)$\\ 
\textbf{Choose:} $a_{ij}$ with~$\sum_ja_{ij}=1$, and $b_{ij}$ with~$\sum_ib_{ij}=1$\\
\For{$k=0,1,\ldots,$}{
\vspace{-0.2cm}
\begin{subequations}\label{ABN}
 	\begin{align}
 	 	&\mbox{\textbf{Transmit:}}~\mb{x}_k^i\mbox{ and }b_{ij}\mb{s}_k^i ~\mbox{to each}~j\in\mc{N}_i^{{\scriptsize \mbox{out}}}\notag\\
 	 	 &\mbox{\textbf{Compute:}}\notag\\
 	 	&\mb{y}^i_{k+1}\leftarrow\textstyle\sum_{j\in\mc{N}_i^{{\tiny \mbox{in}}}}a_{ij}\mb{x}^{j}_k-\alpha\mb{s}^i_k 
 	\label{ABNa}\\
 	&\mb{x}^i_{k+1}\leftarrow \mb{y}^{i}_{k+1} + \beta_k (\mb{y}^i_{k+1}-\mb{y}^i_{k}) \label{ABNb}\\
&\mb{s}^i_{k+1}\leftarrow\textstyle\sum_{j\in\mc{N}_i^{{\tiny \mbox{in}}}}b_{ij}\mb{s}^j_k+\nabla f_i\big(\mb{x}^i_{k+1}\big)-\nabla f_i\big(\mb{x}^i_k\big) \label{ABNd}
 	\end{align}
 \end{subequations}
}
\caption{$\mc{ABN}$}
\label{ABNalg}
\end{algorithm}
\noindent A valid choice for~$b_{ij}$'s at each~$i$ is to choose them as~$1/{|\mc{N}_i^{{\scriptsize \mbox{out}}}|}$, which does not require knowing the outgoing nodes but only the out-degree. For the function class~$\mc{F}^{1,1}_{\mu,L}$,~$\beta$ is a constant; for the function class~$\mc{F}^{1,1}_{L}$, we choose~$\beta_k = \frac{k}{k+3}$.

\subsection{The FROZEN algorithm}
Note that~$\mc{ABN}$ is restricted to communication protocols that allow column-stochastic weights,~$\{b_{ij}\}$'s. When this is not possible, it is desirable to have algorithms that only use row-stochastic weights. Row-stochasticity is trivially established at the receiving agent by assigning a weight to each incoming information such that the sum of weights is~$1$. To avoid CS weights altogether, we now develop a distributed Nesterov gradient method that only row-stochastic weights and show the procedure of constructing this new algorithm from~$\mc{ABN}$. 

To this aim, we first write~$\mc{ABN}$ in the vector-matrix form. Let~$\mb{x}_k,\mb{y}_k$,~$\mb{s}_k$, and~$\nabla\mb f(\mb x_k)$ denote the concatenated vectors with~$\mb{x}_k^i$'s,~$\mb{y}_k^i$'s,~$\mb{s}_k^i$'s, and~$\nabla f_i(\mb x_k^i)$'s, respectively. Then~$\mc{ABN}$ can be compactly written follows:
\begin{subequations}\label{ABNv}
	\begin{align}
	\mb{y}_{k+1}&=\mc{A}\mb{x}_k-\alpha\mb{s}_k, 
	\label{ABNva}\\
	\mb{x}_{k+1}&= \mb{y}_{k+1} + \beta_k (\mb{y}_{k+1}-\mb{y}_{k}), \label{ABNvb}\\
	\mb{s}_{k+1}&=\mc{B}\mb{s}_k+\nabla \mb f\big(\mb{x}_{k+1}\big)-\nabla \mb f\big(\mb{x}_k\big), \label{ABNvc}
	\end{align}
\end{subequations}
where~$\mc{A}=A\otimes I_p$ and~$\mc B=B\otimes I_p$, where~$\otimes$ is the Kronecker. Since~$\mc{A}$ is already row-stochastic, we seek a transformation that makes~$\mc{B}$ a row-stochastic matrix. Since~$B$ is column-stochastic, we denote its left and right Perron eigenvectors as~$\mb{1}_{n}^\top{B}=\mb{1}_{n}^\top$ and~$B\mb{v}=\mb{v}$. Let~$\mbox{diag}(\mb v)$ denote a matrix with~$\mb v$ on its main diagonal. With the help of~$V=\mbox{diag}(\mb{v})\otimes I_p$, we define a state transformation,~$\wt{\mb{s}}_k=V^{-1} \mb{s}_k$, and rewrite~$\mc{ABN}$ as follows:

\newpage
~
\vspace{-.5cm}
\begin{subequations}\label{frost}
	\begin{align}
	\mb{y}_{k+1}&=\mc{A}\mb{x}_k-\alpha V\wt{\mb{s}}_k, 
	\label{frosta}\\
	\mb{x}_{k+1}&= \mb{y}_{k+1} + \beta_k (\mb{y}_{k+1}-\mb{y}_{k}), \label{frostb}\\
	\wt{\mb{s}}_{k+1}&=\wt{\mc{A}}\wt{\mb{s}}_k+V^{-1}\left(\nabla \mb f\big(\mb{x}_{k+1}\big)-\nabla \mb f\big(\mb{x}_k\big)\right), \label{frostc}
	\end{align}
\end{subequations}
where~$\wt{\mc{A}}=V^{-1}\mc{B}V$ can be easily verified to be row-stochastic. Since~$\mb v$ is the right Perron vector of~$\wt{\mc{A}}$, it is not locally known to any agent and thus the above equations are not practically possible to implement. We thus add an independent eigenvector learning algorithm to the above set equations and obtain \textit{FROZEN} (Fast Row-stochastic OptimiZation with Nesterov's momentum) described in Algorithm~\ref{FRalg}. The momentum parameter is chosen the same way as in~$\mc{ABN}$.  

\begin{algorithm}[H]
At each agent~$i$:\\
\textbf{Initialize:} Arbitrary~$\mb{x}_0^i=\mb{y}_0^i\in\mbb{R}^p$,~$\mb{s}_0^i=\nabla f_i(\mb{x}_0^i)$,~$\mb{v}_0^i=\mb{e}_i$\\ 
\textbf{Choose:} $a_{ij}$ with~$\sum_ja_{ij}=1$, and $\wt a_{ij}$ with~$\sum_j\wt a_{ij}=1$\\
\For{$k=1,\ldots,$}{
\vspace{-0.2cm}
 \begin{subequations}\label{fr}
 	\begin{align}
&\mbox{\textbf{Transmit:}}~~\mb{x}_k^i,\mb{v}_k^i,\mb{s}_k^i~\mbox{to each}~j\in\mc{N}_i^{{\scriptsize \mbox{out}}}\notag\\
 	 	 &\mbox{\textbf{Compute:}}\notag\\ 	
 	 	 &\mb{v}^i_{k+1} \leftarrow \textstyle\sum_{j\in\mc{N}_i^{{\tiny \mbox{in}}}} \wt a_{ij}\mb{v}_k^j\\
 	&\mb{y}^i_{k+1} \leftarrow \textstyle\sum_{j\in\mc{N}_i^{{\tiny \mbox{in}}}}a_{ij}\mb{x}^{j}_k-\alpha {\mb{s}}^i_k 
 	\label{fra}\\
 	&\mb{x}^i_{k+1} \leftarrow \mb{y}^{i}_{k+1} + \beta_k (\mb{y}^i_{k+1}-\mb{y}^i_{k}) \label{frb}\\
 	&{\mb{s}}^i_{k+1} \leftarrow \textstyle\sum_{j\in\mc{N}_i^{{\tiny \mbox{in}}}}\wt a_{ij}\mb{s}^j_k+\frac{\nabla f_i\left(\mb{x}^i_{k+1}\right)}{[\mb{v}^i_{k+1}]_i}-\frac{\nabla f_i\left(\mb{x}^i_k\right)}{[\mb{v}^i_{k}]_i} \label{frc}
 	\end{align}
 \end{subequations}
}
\caption{\textit{FROZEN}}
\label{FRalg}
\end{algorithm}
In the above algorithm,~$\mb{e}_0^i\in\mbb{R}^n$ is a vector of zeros with a~$1$ at the~$i$th location and~$[\:\cdot\:]_i$ denotes the~$i$th element of a vector. We note that although the weight assignment in \textit{FROZEN} is straightforward, this flexibility comes at a price: 
\begin{inparaenum}[(i)]
\item each agent must maintain an additional~$n$-dimensional vector,~$\mb{v}_k^i$;
\item additional iterations are required for eigenvector learning in Eq.~\eqref{fra}; and, 
\item the initial condition~$\mb{v}_0^i=\mb{e}_0^i$ requires each agent to have and know a unique identifier. 
\end{inparaenum}
However, as discussed earlier,~$\mc{ABN}$ may not be applicable in some communication protocols and thus, \textit{FROZEN} may be the only algorithm available. Finally, we note that when~$\beta_k=0,\forall k$, \textit{FROZEN} reduces to FROST whose detailed analysis and a linear convergence proof can be found in~\cite{linear_row,FROST}.

\textbf{Generalizations and extensions: }The method we described to convert~$\mc{ABN}$ to \textit{FROZEN} leads to another variant of~$\mc{ABN}$ with only CS weights, see~\cite{AB} for details. The resulting methods add Nesterov's momentum to ADDOPT and Push-DIGing~\cite{add-opt,diging}. Since these variants only require CS weights,~$\mc{AB}$ and~$\mc{ABN}$ are preferable due to their faster convergence. It is further straightforward to conceive a time-varying implementation of~$\mc{ABN}$ and~\textit{FROZEN} over gossip based protocols or random graphs, see e.g., the related work in~\cite{TVAB,TVpushpull} on non-accelerated methods. Asynchronous schemes may also be derived following the methodologies studied in~\cite{Async_PG_EXTRA,ASY-SONATA}. Finally, we note that a rigorous theoretical analysis of~$\mc{AB}$ and~$\mc{ABN}$ is beyond the scope of this letter. We thus rely on simulations to highlight and verify different aspects of the proposed methods.

\newpage
\section{Numerical Results}\label{s4}
In this section, we numerically verify the convergence of the proposed algorithms,~$\mc{ABN}$ and \textit{FROZEN}, in this letter, and compare them with well-known solutions for distributed optimization. To this aim, we generate strongly-connected digraphs with~$n=30$ nodes using nearest-neighbor rules. We use an uniform weighting strategy to generate the row- and column-stochastic weight matrices, i.e.,~$a_{ij} = 1/|\mc{N}_i^{{\scriptsize \mbox{in}}}|, \forall i,$ and~$b_{ij} = 1/|\mc{N}_j^{{\scriptsize \mbox{out}}}|, \forall j$. We first compare~$\mc{ABN}$ and \textit{FROZEN} with the following methods over digraphs: ADDOPT/Push-DIGing~\cite{add-opt,diging}, FROST~\cite{FROST}, and~$\mc{AB}$~\cite{AB}. For comparison, we plot the average residual:~$\frac{1}{n}\sum_{i=1}^{n}\|\mb{x}_i(k)-\mb{x}^*\|_2$. 

\vspace{-0.15cm}
\subsection{Strongly-convex case}
\begin{figure}[!h]
	\centering
\includegraphics[width=2.5in]{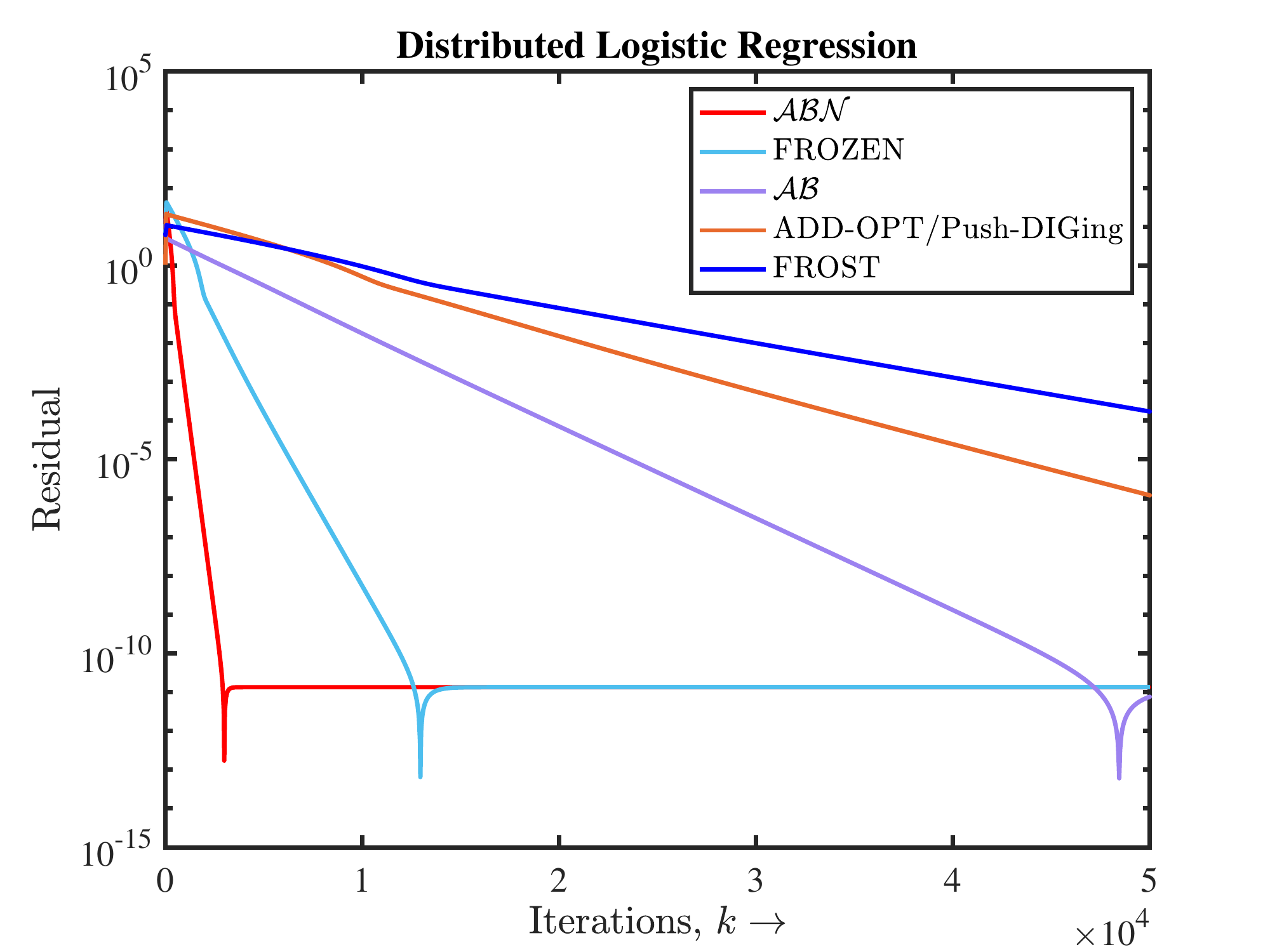}
	\caption{Strongly-convex case: Accelerated linear rate}
	\label{SC}
\end{figure}

We first consider a distributed binary classification problem using logistic loss: each agent~$i$ has access to~$m_i$ training samples,~$(\mb{c}_{ij},y_{ij})\in\mathbb{R}^p\times\{-1,+1\}$, where~$\mb{c}_{ij}$ contains~$p$ features of the~$j$th training data at agent~$i$, and~$y_{ij}$ is the corresponding binary label. The agents cooperatively minimize~$F=\sum_{i=1}^nf_i(\mb{b},c)$, where~$\mb{b}\in\mbb{R}^p,c\in\mbb{R}$ are the optimization variables to learn the separating hyperplane, with each~$f_i$ being
\begin{equation}
	f_i(\mb{b},c)=\textstyle\sum_{j=1}^{m_i}\ln[1+e^{-(\mb{b}^\top\mb{c}_{ij}+c)y_{ij}}]+\frac{\lambda}{2}(\|\mb{b}\|_2^2+c^2). \nonumber
	\end{equation}
In our setting, the feature vectors,~$\mb{c}_{ij}$'s, are  generated from a Gaussian distribution with zero mean. The binary labels are generated from a Bernoulli distribution. We set~$p = 10$ and~$m_i = 5, \forall i$. The results are shown in Fig.~\ref{SC}. Although \textit{FROZEN} is slower than~$\mc{ABN}$, it is applicable broadcast-based protocols as it only requires row-stochastic weights. The step-size and momentum parameters are manually chosen to obtain the best performance for each algorithm.
 
\vspace{-0.35cm}
\subsection{Non strongly-convex case}
\begin{figure}[!h]
	\centering
	\includegraphics[width=2.5in]{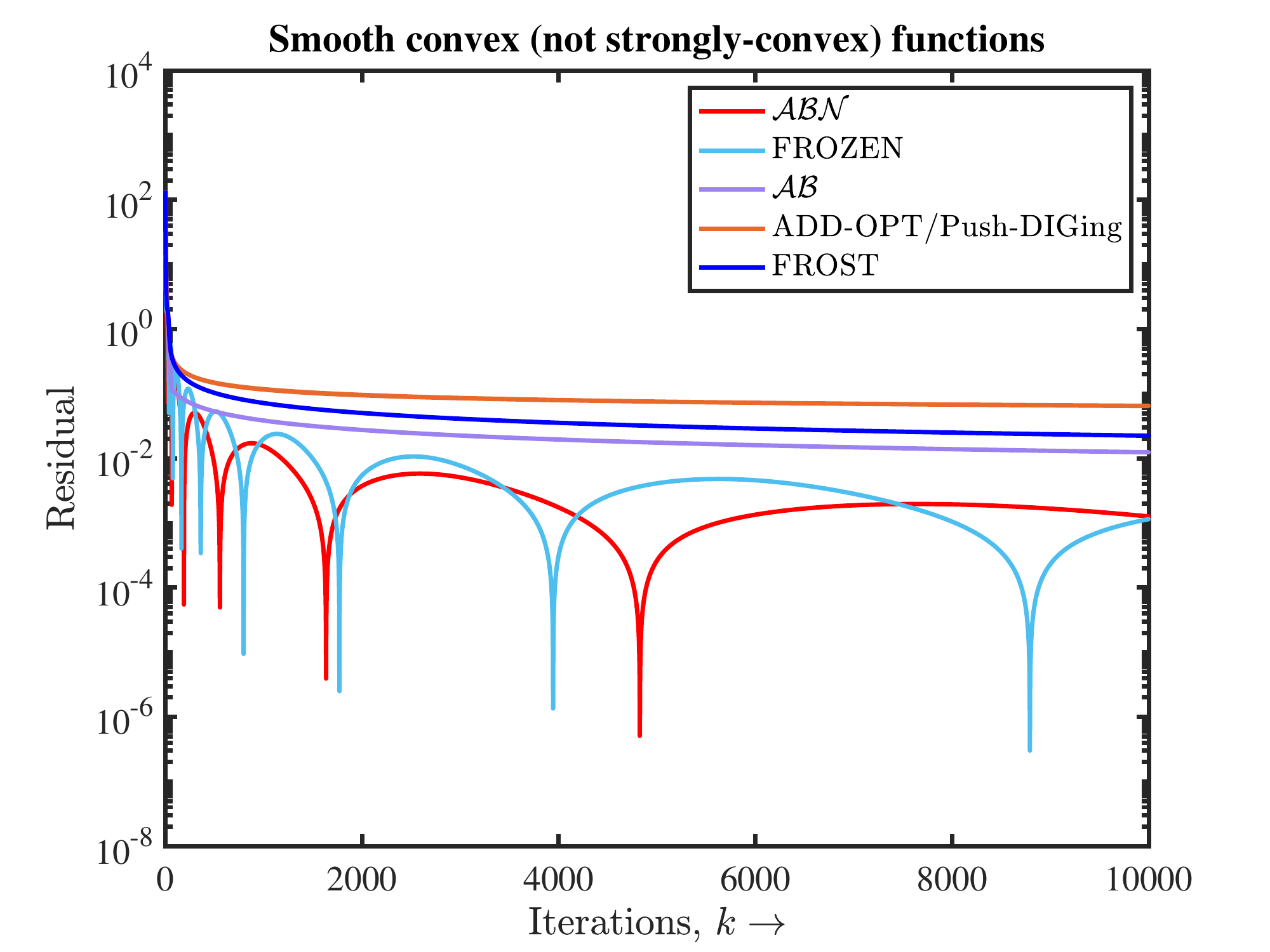} 
	\caption{Non strongly-convex case: Accelerated sublinear rate}
	\label{NSC}
\vspace{-0.25cm}
\end{figure}

We next choose the objective functions,~$f_i$'s, to be smooth, convex but not strongly-convex. In particular,~$f_i(x) = u(x)+b_i x$, where~$b_i$'s are randomly generated,~$b_n = -\sum_{i=1}^{n-1}b_i$, and~$u(x)$ is chosen as follows:
\begin{align*} 
u(x)=\left\{
\begin{array}{rl}
\frac{1}{4}x^4,&|x|\leq 1,\\
|x|-\frac{3}{4},&|x|> 1.
\end{array}
\right.
\end{align*} 
It can be verified that~$f = \sum_{i}f_i$ is not strongly-convex as~$f^{''}(x^*)=0$. The results are shown in Fig.~\ref{NSC} where the momentum parameter is chosen as~$\beta_k=\frac{k}{k+3}$ and other parameters are manually optimized.

\vspace{-.4cm}
\subsection{Influence of graph sparsity}
\begin{figure}[!h]
	\centering
	\subfigure{\includegraphics[width=1.13in]{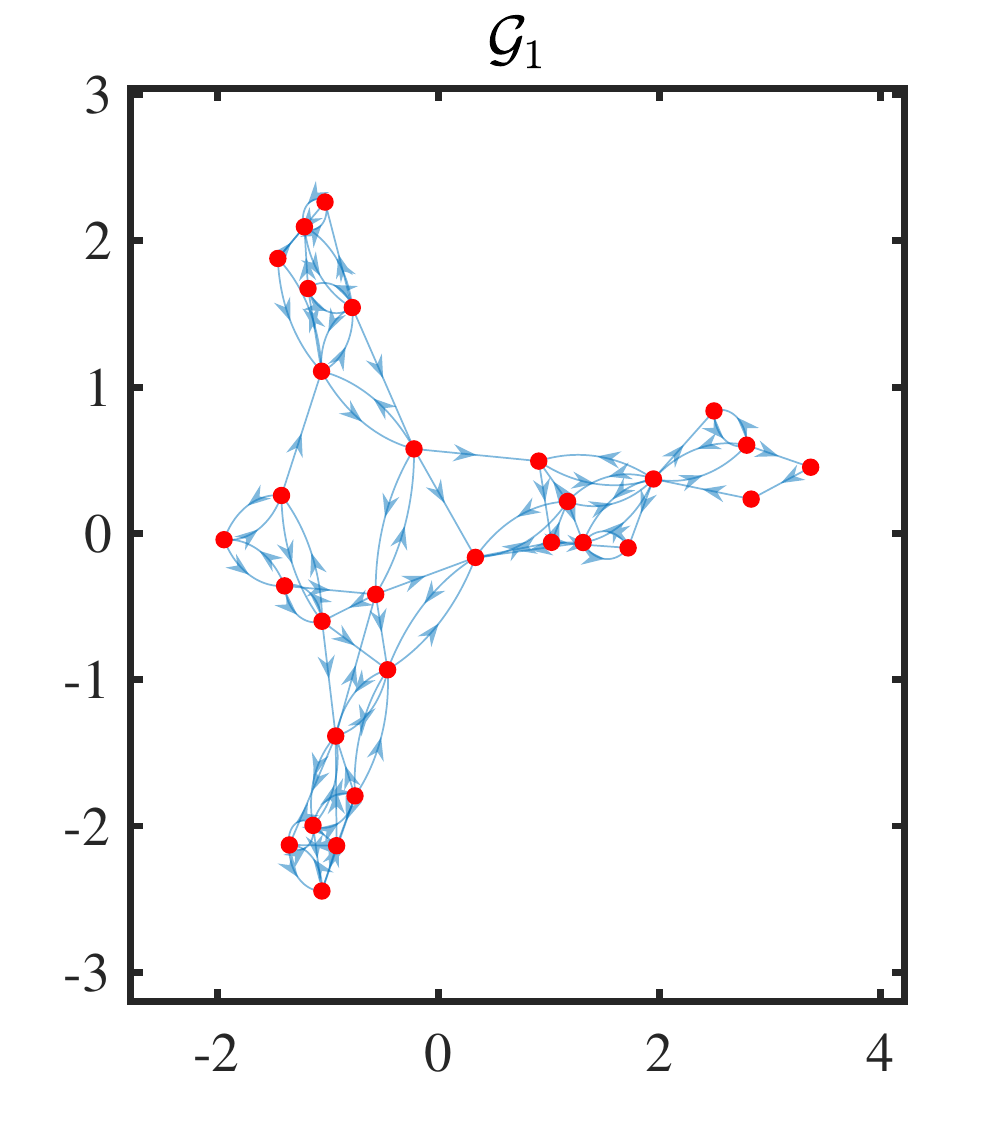}}
	\subfigure{\includegraphics[width=1.13in]{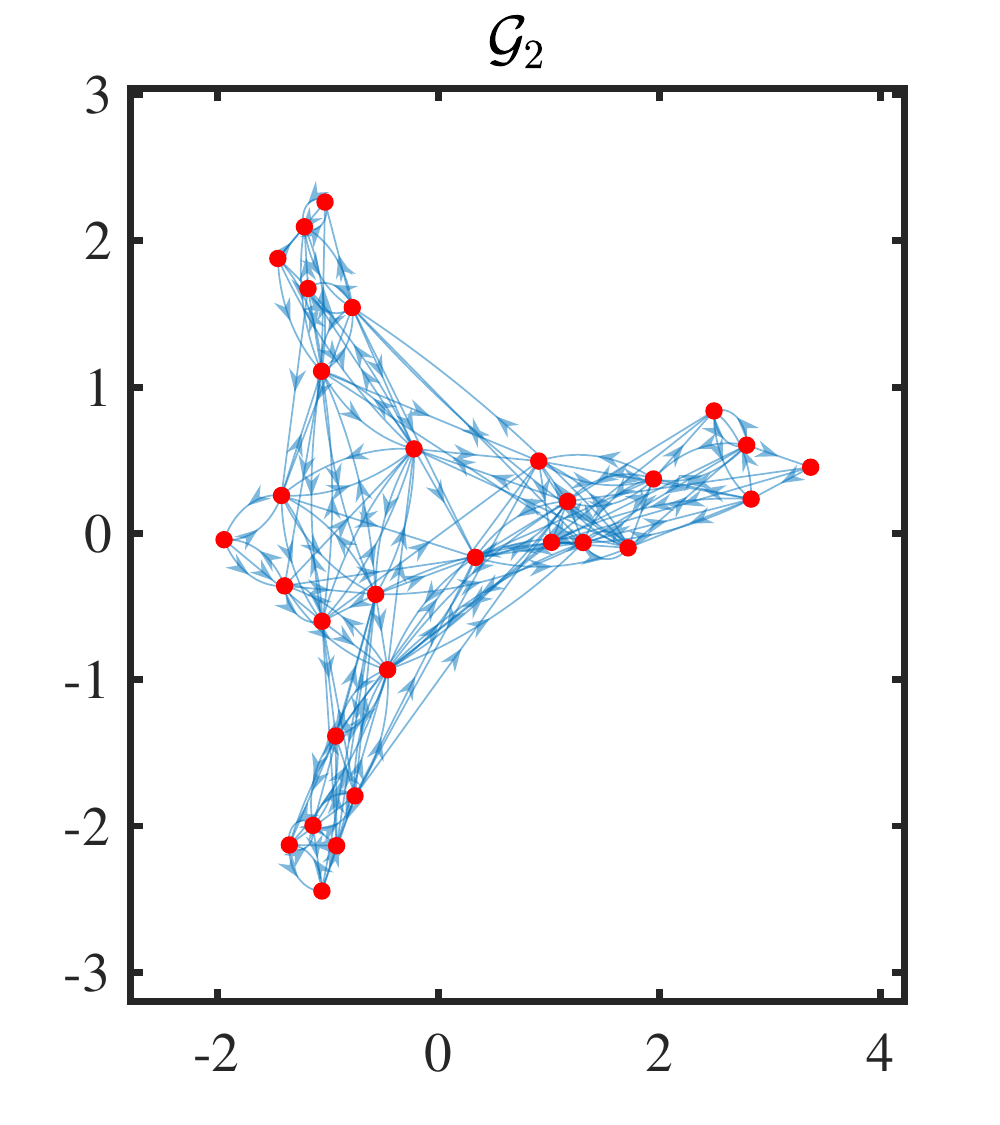}}
	\subfigure{\includegraphics[width=1.13in]{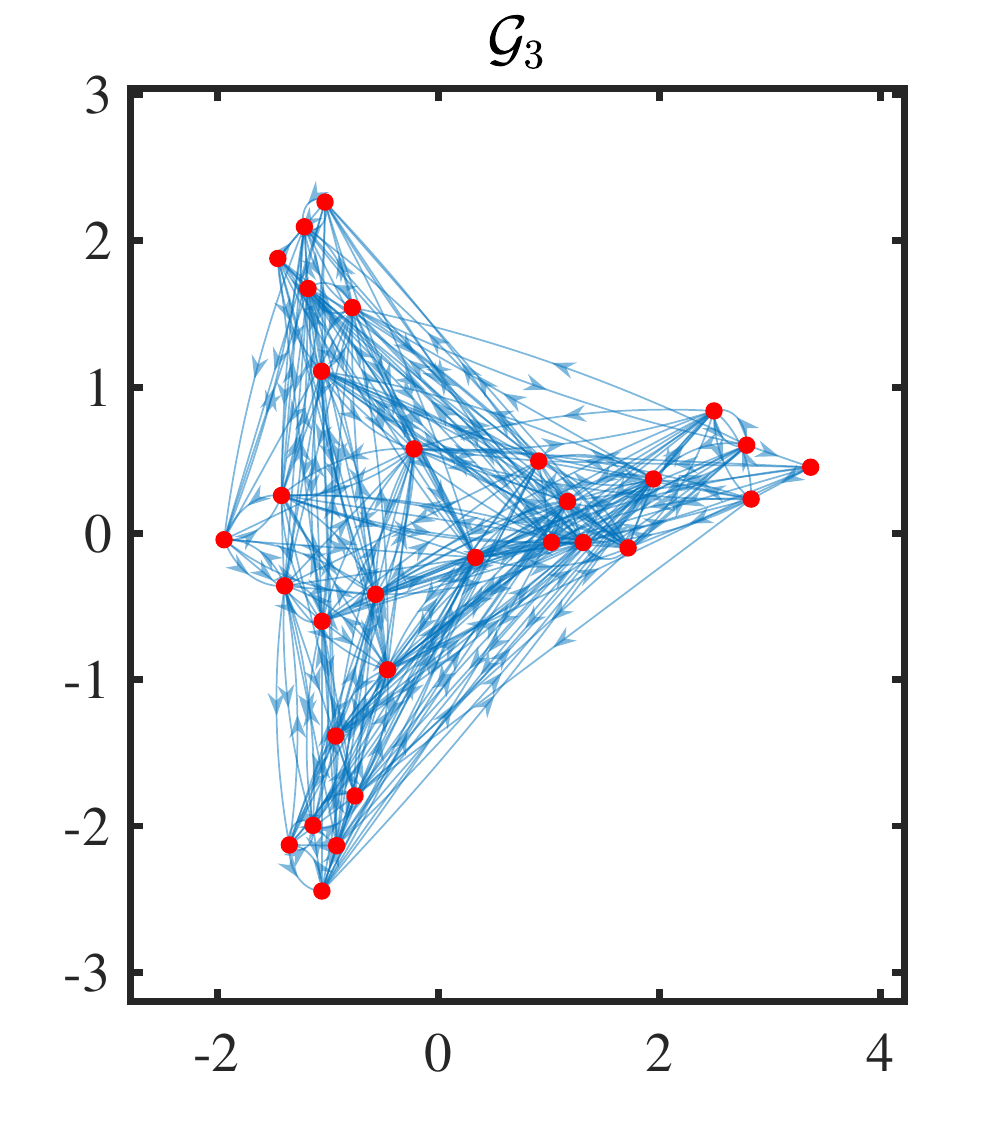}}
	\subfigure{\includegraphics[width=1.7in]{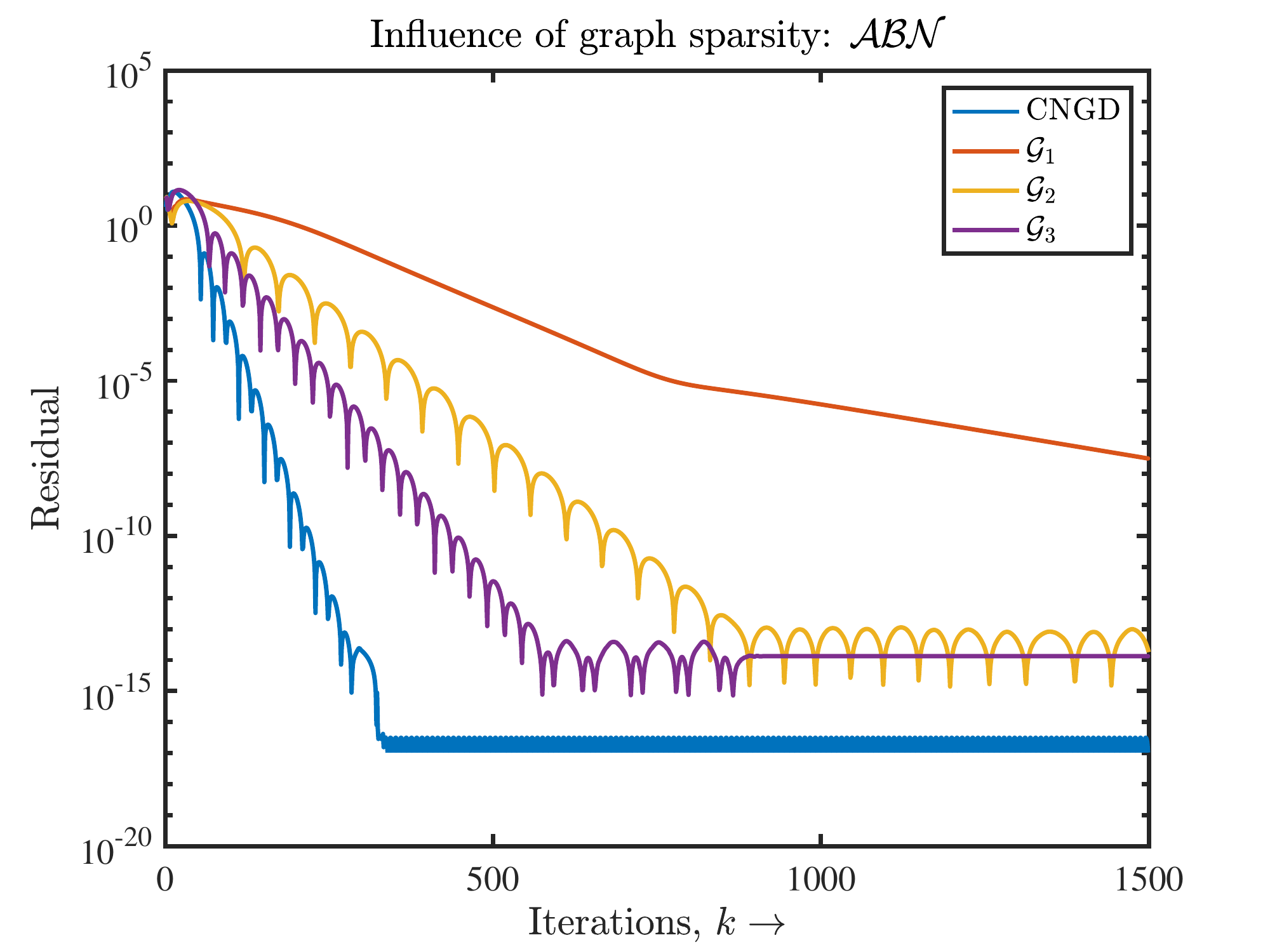}}
	\subfigure{\includegraphics[width=1.7in]{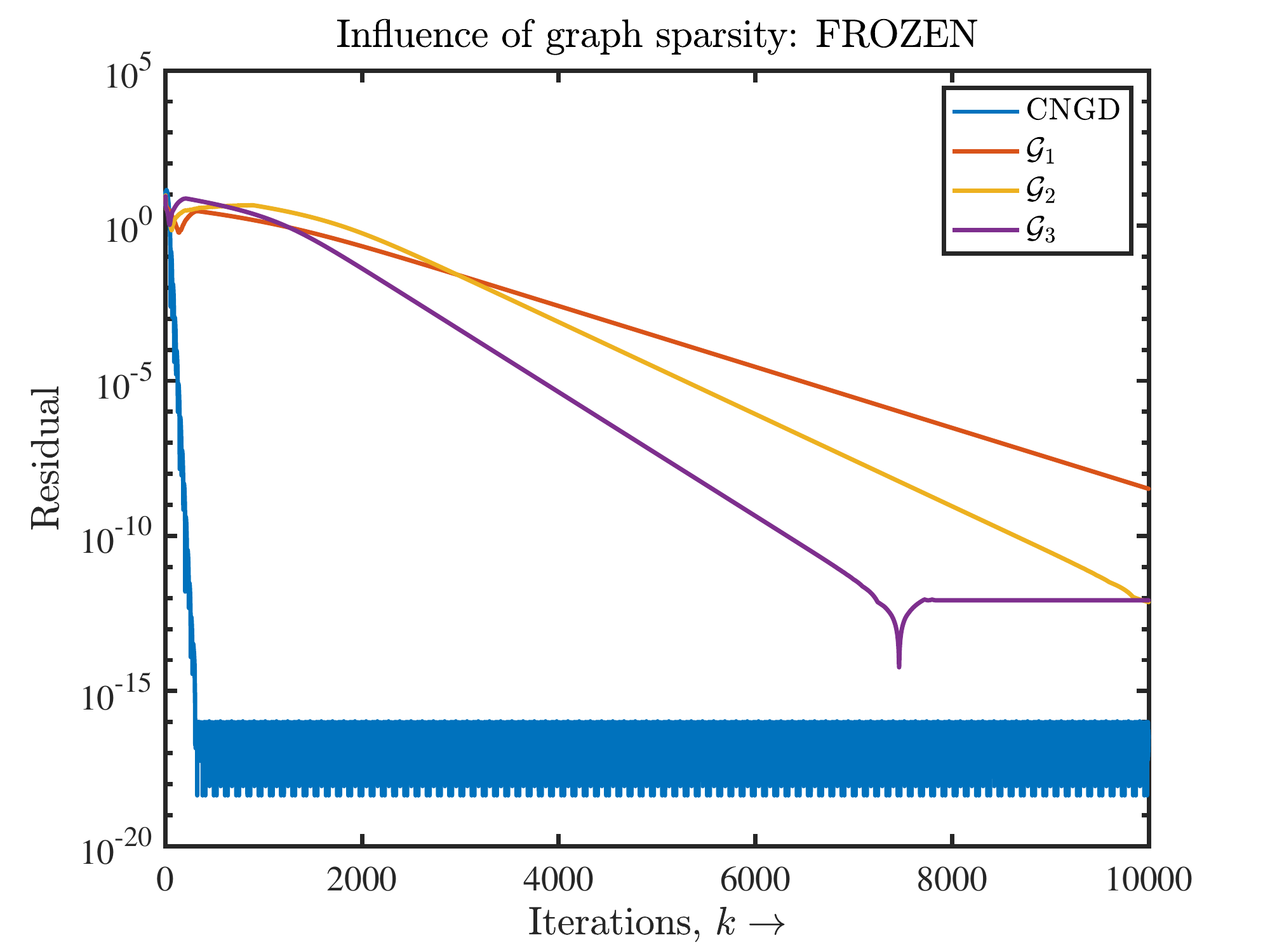}}
	\caption{Influence of digraph sparsity on~$\mc{ABN}$ and \textit{FROZEN}.}
	\label{spar}
\end{figure}

Finally, we study the influence of graph sparsity with the help of the logistic regression problem discussed earlier. We fix the number of nodes to~$n=30$ and randomly generate three nearest-neighbor digraphs,~$\mc{G}_1$,~$\mc{G}_2$ and~$\mc{G}_3$, with decreasing sparsity, see Fig.~\ref{spar} (Top). In Fig.~\ref{spar} (Bottom), we compare the performance of the proposed methods with centralized Nesterov over the three graphs. It can be verified that~$\mc{ABN}$ and \textit{FROZEN} approach centralized Nesterov method as the graphs become dense. \textit{FROZEN}, however, is much slower than~$\mc{ABN}$ because it additionally requires eigenvector learning. 

\vspace{-0.25cm}
\section{Conclusions}\label{s5}
In this letter, we present accelerated methods for optimization based on Nesterov's momentum over arbitrary, strongly-connected, graphs. The fundamental algorithm,~$\mc{ABN}$, uses both row- and column-stochastic weights, simultaneously, to achieve agreement and optimality. We then derive a variant from~$\mc{ABN}$, termed as~\textit{FROZEN}, that only uses row-stochastic weights and thus is applicable to a larger set of communication protocols, however, at the expense of eigenvector learning, thus resulting into slower convergence. Although a theoretical analysis is beyond the scope of this letter, we provide an extensive set of numerical results to study the behavior of the proposed methods for both convex and strongly-convex cases.

\bibliographystyle{IEEEbib}
\bibliography{sample1,ref}

\end{document}